%% file: main.tex
\documentclass[conference]{IEEEtran}
\IEEEoverridecommandlockouts
\usepackage{cite}
\usepackage{amsmath,amssymb,amsfonts}
\usepackage{algorithmic}
\usepackage{graphicx}
\usepackage{textcomp}
\usepackage{cite}
\usepackage{booktabs}
\usepackage{multirow}
\usepackage{xcolor}
\def\BibTeX{{\rm B\kern-.05em{\sc i\kern-.025em b}\kern-.08em
    T\kern-.1667em\lower.7ex\hbox{E}\kern-.125emX}}
\begin{document}

\title{DiffETM: \textbf{Diff}usion Process Enhanced \textbf{E}mbedded \textbf{T}opic \textbf{M}odel
\thanks{* Corresponding Author}
}

\author{\IEEEauthorblockN{1\textsuperscript{st} Wei Shao}
\IEEEauthorblockA{\textit{Department of Computer Science} \\
\textit{City University of Hong Kong}\\
Hong Kong, Hong Kong SAR \\
weishao4-c@my.cityu.edu.hk}
\and
\IEEEauthorblockN{2\textsuperscript{nd} Mingyang Liu}
\IEEEauthorblockA{\textit{Department of Computer Science} \\
\textit{City University of Hong Kong}\\
Hong Kong, Hong Kong SAR \\
mingyaliu8-c@my.cityu.edu.hk}
\and
\IEEEauthorblockN{3\textsuperscript{rd} Linqi Song*}
\IEEEauthorblockA{\textit{Department of Computer Science} \\
\textit{City University of Hong Kong}\\
Hong Kong, Hong Kong SAR \\
song.linqi@cityu.edu.hk}
}

\maketitle

\begin{abstract}

The embedded topic model (ETM) is a widely used approach that assumes the sampled document-topic distribution conforms to the logistic normal distribution for easier optimization. However, this assumption oversimplifies the real document-topic distribution, limiting the model's performance. In response, we propose a novel method that introduces the diffusion process into the sampling process of document-topic distribution to overcome this limitation and maintain an easy optimization process. We validate our method through extensive experiments on two mainstream datasets, proving its effectiveness in improving topic modeling performance.
\end{abstract}

\begin{IEEEkeywords}
topic modeling, embedded topic model, diffusion process, text mining
\end{IEEEkeywords}

\input{sections/introduction}
\input{sections/prelim}
\input{sections/method}

\input{sections/experiment}
\input{sections/conclusion}

\section{Acknowledgement}
This work was supported in part by the National Natural Science Foundation of China under Grant 62371411, the Research Grants Council of the Hong Kong SAR under Grant GRF 11217823 and Collaborative Research Fund C1042-23GF, InnoHK initiative, the Government of the HKSAR, Laboratory for AI-Powered Financial Technologies.

\bibliography{main}
\bibliographystyle{IEEEtran}


\end{document}

%% file: sections/introduction.tex
\section{Introduction}
\begin{figure}
    \centering
    \includegraphics[width=0.9\linewidth]{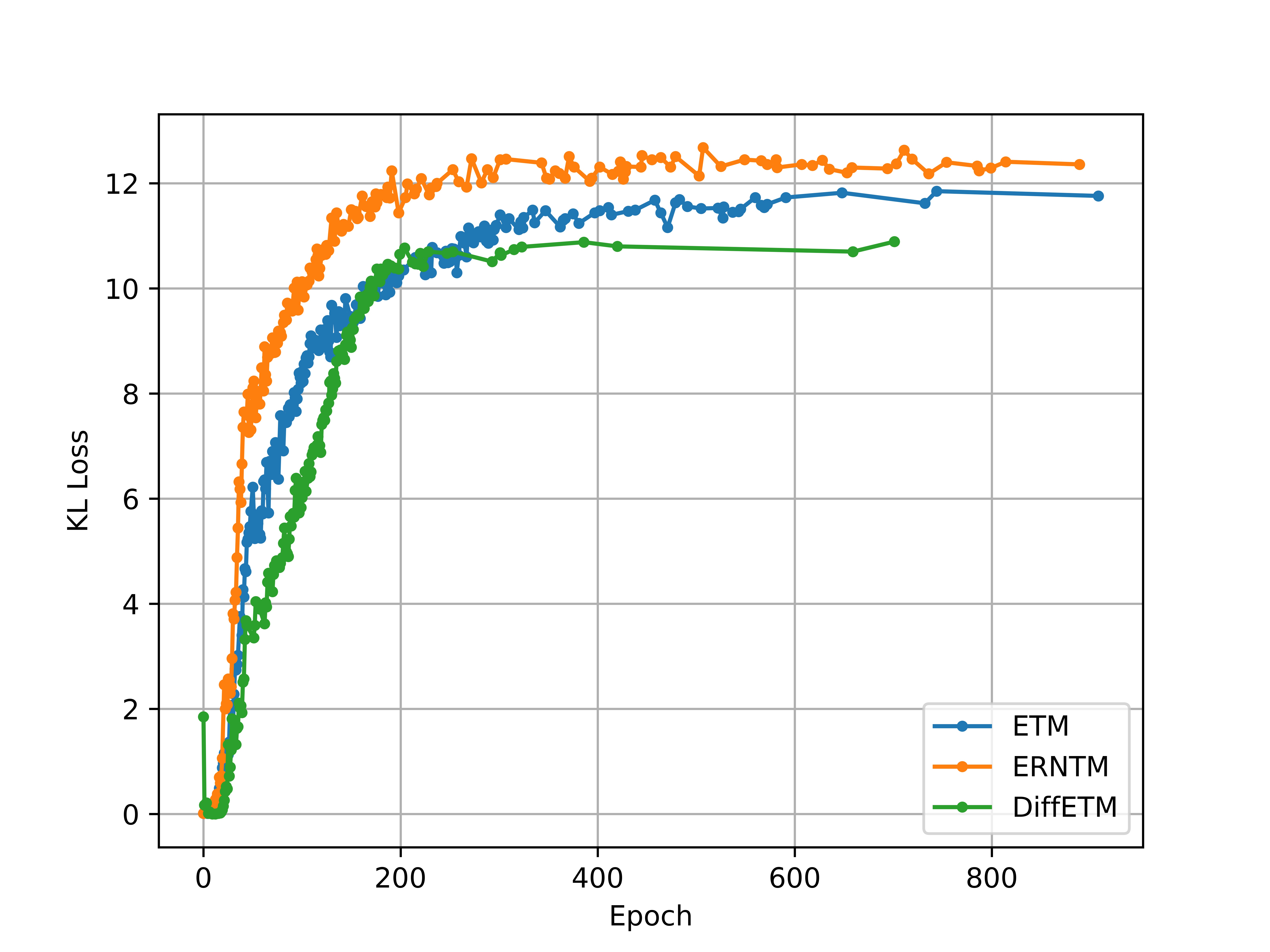}
    \caption{This figure shows the KL loss changes of several classical embedded topic models on the test set of the 20NewsGroup dataset when the topic number is 50. Each point means a better model checkpoint than previous ones. The loss is the KL divergence between sampled topic distribution variables and the normal distribution. According to this figure, we could find that the loss keeps being larger in the training process. This demonstrates that when the sampled document-topic distributions tend to break the limitations of conforming to the logistic-normal distribution for a better topic modeling performance.}
    \label{KLchange}
\end{figure}

In recent years, the embedded topic model~\cite{dieng2020topic} has garnered significant attention~\cite{Zhang2020CombineTM, Akash2022CoordinatedTM, Shao2022TowardsBU, Seifollahi2021AnET, wu2023effective,xu-etal-2023-detime,chen-etal-2023-nonlinear,NEURIPS2023_fce17645} due to its interpretable and flexible variational auto-encoder architecture~\cite{Kingma2013AutoEncodingVB,miao2017discovering}. Despite its merits, the embedded topic model still faces a critical challenge: it assumes that the topic distribution of a document assumes to the logistic-normal distribution and thus employs a simple yet effective variational loss for training. While this assumption facilitates easier optimization, it also imposes a strict constraint on the learned document-topic distribution. As a result, the embedded topic model struggles to achieve higher performance in topic modeling, as it fails to capture the complexity of the real document-topic distribution (we further illustrate this limitation in Fig.~\ref{KLchange}).

To alleviate this problem, we rethink the architecture of the embedded topic model, which is a variational auto-encoder. Compared to the standard auto-encoder~\cite{bank2021autoencoders}, it tries to model more complex document-topic distribution by introducing noise to the hidden representations sampled from a standard normal distribution. The noise is fused into these hidden representations in the form of mean and variance and thus results in a variable following a Gaussian distribution. This variable is then transformed into document-topic distributions using a softmax function. However, according to the experimental results we observed, the introduction of these noises is still insufficient to model the document topic distribution well.

To this end, we propose a new idea of representation enhancement: we directly sample these hidden representations from document representations. 
The advantage of this idea is that compared to ETM, we integrate document information into these hidden representations, improving its ability to model document-topic distribution. On the other hand, the disadvantage is that due to the change in the hidden representation's distribution, we cannot use the existing objective function for optimization, and it is difficult for us to obtain an accurate loss function.

To alleviate this disadvantage, we combine the sampling process with the forward process of diffusion model~\cite{Nichol2021ImprovedDD}, which gradually introduces noise conforming to the normal distribution into the document representations, following the steps of the diffusion process. Finally, the resulting representation contains both document information and is close enough to the normal distribution and thus we can still utilize the objective function of the embedded topic model to optimize the new model.

By utilizing this diffusion-based approach, our proposed model achieves significant improvements across three metrics: topic coherence, topic diversity, and perplexity, for topic modeling on two widely-used datasets, namely 20Newsgroup~\cite{Newsgroups20} and the New York Times~\cite{10.1007/s11042-018-6894-4}. When compared to both classical and state-of-the-art embedded topic models, our model demonstrates significant improvement in performance. As far as we know, we are the first to introduce the diffusion process into the embedded topic model to enhance the representation ability of the document-topic distribution.

%% file: sections/prelim.tex
\section{Preliminary}
Topic modeling is a task to obtain the hidden document-topic distribution via modeling the whole document set. For a set with $N$ documents, it has $V$ unique words. We use a bag-of-words model~\cite{doi:10.1080/00437956.1954.11659520} to represent each document $\boldsymbol{X_i} \in \mathbb{R}^{V}$.  There is a latent topic set $\boldsymbol{Z}=\{z_1, z_2, ..., z_K\}$ consisting of $K$ latent topics in the document set and each document $\boldsymbol{X_i}$ has a distribution $\boldsymbol{\theta_i} \in \mathbb{R}^{1 \times K}$ over this topic set (document-topic distribution). For each topic $z_i$, there is also a distribution $\boldsymbol{\beta_i} \in \mathbb{R}^{1 \times V}$ over vocabulary. 

The topic model aims to model the document set and the modeling process is equivalent to maximizing the likelihood of documents:
\begin{equation}
    L = \sum\nolimits^{N}_{i=1}\log p(\boldsymbol{X_i})\,, 
\end{equation}
\begin{equation}
    p(\boldsymbol{X_i}) = \prod^{V}_{j=1} (\sum^{K}_{k=1}p(z_k|\boldsymbol{X_i})p(w_j|z_k))^{X_{ij}}\,,
\end{equation}
\begin{equation}
    p(\boldsymbol{X_i}) = \prod\nolimits^{V}_{j=1} (\boldsymbol{\theta_{i}} \times \boldsymbol{\beta})^{X_{ij}}\,.
\end{equation}

where $w_j$ is the j-th word in the vocabulary, and the $X_{ij}$ means the number of occurrences of $w_j$ in $X_i$. In the embedded topic model, words and topics are projected into a vector space and form two embedding matrices: word embedding matrix $\boldsymbol{\rho} \in R^{V \times E}$ and topic embedding matrix $\boldsymbol{\alpha} \in R^{K \times E}$. The topic-word distribution $\boldsymbol{\beta} = softmax(\boldsymbol{\alpha} \times \boldsymbol{\rho})$.

%% file: sections/method.tex
\section{Methodology}
\begin{figure*}
    \centering
    \includegraphics[width=0.9\linewidth]{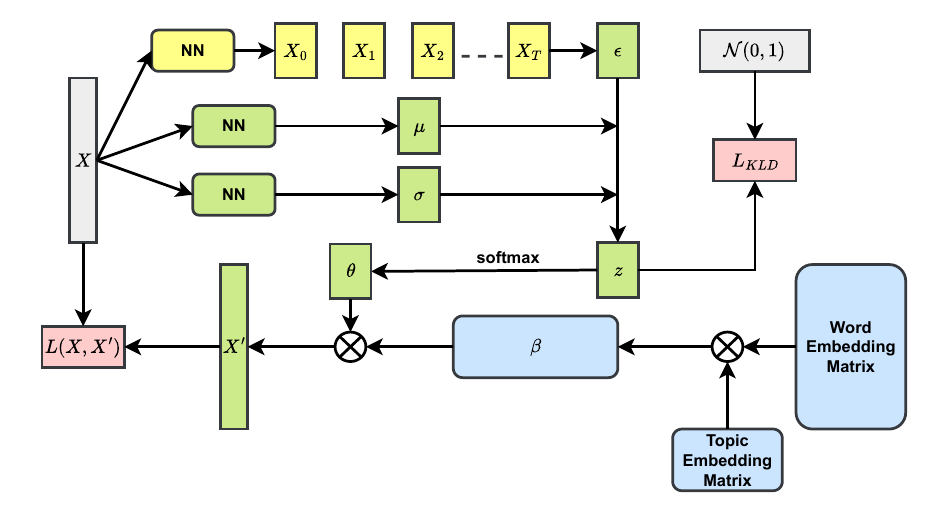}
    \caption{This is the architecture of our proposed model, including the diffusion module (in yellow), document-topic distribution computation module (in green), and topic-word distribution module (in blue).}
    \label{model}
\end{figure*}
\subsection{Overview}
The architecture of our model is shown in Fig.~\ref{model}, which includes three modules: diffusion module (in yellow), document-topic distribution computation module (in green), and topic-word distribution computation module (in blue). 

The diffusion module and document-topic distribution module work together to produce the document-topic distribution $\boldsymbol{\theta}$, and the topic-word distribution module could generate the topic-word distribution $\boldsymbol{\beta}$. Then, the two distributions' product $\boldsymbol{X'}$ is regarded as the reconstruction of the input document's normalized bag-of-words representation $\boldsymbol{X}$. In the following sections, we will introduce each module and how to train the whole model in detail.

\subsection{Diffusion Process}
At first, we use a feed-forward network, consisting of three linear layers and two ReLU activation functions, to produce the enhanced representation $\boldsymbol{X_0}$ of document representation $\boldsymbol{X}$.

\begin{equation}
    \boldsymbol{X_0} = NN(\boldsymbol{X}).
\end{equation}

The $\boldsymbol{X_0}$ is the input of the following diffusion process.
The diffusion model's forward process adds noise to the inputs at different levels step by step. There are various methods to add noise. To ensure the simplicity of the topic model, we utilize the linear noise scheduler in our model. The linear noise scheduler adds the noise to the input document $\boldsymbol{X_0}$ and finally produces $\boldsymbol{X_T}$ as the hidden representations $\boldsymbol \epsilon$, which is very close to the normal distribution. We could represent the process as:

$$ q(\boldsymbol{X_T}|\boldsymbol{X_0}) = \mathcal N(X_T; \sqrt{\overline{\alpha}_T} \boldsymbol{X_0}, (1 - \overline{\alpha}_T) I ),$$ 

where $\overline{\alpha}_T = \prod^T_{s=1}\alpha_T, a_T = 1 - \beta_T$. In the embedded topic model, the initial value $\epsilon$ is usually sampled from the normal distribution $\mathcal N(0, 1)$.

\subsection{Document-Topic Distribution Module}
In this module, the noise is added to $\epsilon$ in the form of mean and variance. The mean $\mu$ and variance $\sigma$ are obtained from two neural network layers with the same architecture as the neural network module in the diffusion module. Then, $\epsilon$ is multiplied by the standard deviation $\sigma$, added to $\mu$, and finally passed through a softmax function to produce the document-topic distribution $\theta$. It is represented by:

\begin{equation}
   \boldsymbol{\mu} = NN(\boldsymbol{X};\boldsymbol{v_{\mu}})\,,
\end{equation}
\begin{equation}
    \boldsymbol{\sigma} = NN(\boldsymbol{X};\boldsymbol{v_{\sigma}})\,,
\end{equation}
\begin{equation}
    \boldsymbol{z} = \boldsymbol{\epsilon} \odot \boldsymbol{\sigma} + \boldsymbol{\mu}\\,
\end{equation}
\begin{equation}
    \boldsymbol{\theta} = softmax(\boldsymbol{z})\,,
\end{equation}
where the $\boldsymbol{v_{\mu}}$ and $\boldsymbol{v_{\sigma}}$ are parameters of the neural networks.

\subsection{Topic-Word Distribution Module}
In the embedded topic model, the topic-word distribution is the product of topic embedding and word embedding:
\begin{equation}
    \boldsymbol{\beta} = \boldsymbol{\alpha} \times \boldsymbol{\rho}^\top\,.
\end{equation}

Each element in vector $\boldsymbol{\beta}_{k}$ epresents the probability of the corresponding word belonging to topic $k$ (e.g., $\boldsymbol{\beta}_{i,j} = p(w_j|z_k)$).
\subsection{Training}
This model is trained via two loss terms: one is the reconstruction loss $L(\boldsymbol{X}, \boldsymbol{X'})$:
\begin{equation}
    \boldsymbol{L(X, X')} = \boldsymbol{X} \text{log}\boldsymbol{X'}\,,
\end{equation}
And another term is the KL divergence between the produced $z$ and the normal distribution $\mathcal N(0, 1)$:
\begin{equation}
    \boldsymbol{L_{KLD}} = KL(\boldsymbol{z}||\mathcal N(0, 1))\,.
\end{equation}

The final loss function is:
\begin{equation}
    \boldsymbol{L} = \boldsymbol{L(X, X')} + \lambda * \boldsymbol{L_{KLD}}\,,
\end{equation} where $\lambda$ is a hyperparameter used to balance the trade-off between the two terms. The whole parameters include the neural networks' parameters in diffusion and document-topic distribution module and the topic and word embedding matrices.

%% file: sections/experiment.tex
\begin{table}[!htb]
    \centering
    
    \scalebox{0.9}{
    \begin{tabular}[width=\textwidth]{|c|c|c|c|}
        \hline
        DATASET & Partition & \ Document Count & \ Vocabulary Size \\
        \hline
        \multirow{3}{*}{20NewsGroup} & train & 10132 &  \\
        \cline{2-3}
        & valid & 1126 & 1994 \\
        \cline{2-3}
        & test & 7487 & \\
        \hline
        \multirow{3}{*}{NYT-10000} & train & 	254616 &  \\
        \cline{2-3}
        & valid & 14978 & 1483 \\
        \cline{2-3}
        & test & 29934 &\\
        \hline
        \multirow{3}{*}{NYT-5000} & train & 254666 &  \\
        \cline{2-3}
        & valid & 14982 & 2889 \\
        \cline{2-3}
        & test & 29947 &\\
        \hline
        \multirow{3}{*}{NYT-3000} & train & 254671 &  \\
        \cline{2-3}
        & valid & 14982 & 4324 \\
        \cline{2-3}
        & test & 29952 &\\
        \hline
    \end{tabular}
    }
    \vspace{0.5em}
    \caption{Statistical information and data partition of datasets used in this paper. NYT-x indicates removing words whose document frequency is less than x}
    \label{tab:data}
\end{table}{}

\linespread{1.25}
\begin{table*}[!ht]
    \centering
    \resizebox{0.9\textwidth}{20mm}{
    \begin{tabular}{c|cccc|cccc|cccc}
    \toprule
    \multirow{3}{*}{METHODS} & \multicolumn{12}{c}{Topic Number} \\
    \cline{2-13} 
    & \multicolumn{4}{c|}{K=50} & \multicolumn{4}{c|}{K=100} & \multicolumn{4}{c}{K=200}   \\
	& Coherence &	Diversity &  Quality & Perplexity & Coherence &	Diversity &  Quality & Perplexity  &  Coherence &	Diversity &  Quality & Perplexity  	\\

\midrule
NTMR & 0.1035 & 0.0776 & 0.0080 & 1135.1 & 0.1123 & 0.0472 & 0.0053 & 1060.7 & 0.1324 & 0.0126 & 0.0017 & 1174.8    \\

NTM	&	0.1804 & 0.3648	&	0.0658 & 897.4	&	0.1621 & 0.2592	&	0.0420 & 852.7	& 0.1037 & 0.1754	&	0.0182 & 823.8 \\


ETM	&	0.1865 & 0.4864	&	0.0907 & 686.0	&	0.1821 & 0.3552	&	0.0647 & 660.0	&	0.1826 &  0.2326	&	0.0425 & 681.0\\

ERNTM	&	0.1949 & 0.5112	& 0.0996 & 651.1	&	0.1873 & 0.3624 &0.0679 & 653.9	&0.1867 & 0.2360	&	0.0441 & 671.2     \\

DeTiME	&	0.1935 & 0.5203	& 0.1007 & 613.4 &	0.1890 & 0.3834 & 0.0725 &	619.1 & 0.1862 & 0.2689	&	0.0501 & 625.6     \\

Meta-CETM	&	0.1958 & 0.4538	& 0.0889 & 591.3	&	0.1901 & 0.3106 & 0.0590 & 553.9	& 0.1845 & 0.2177	& 0.0402 & 601.5     \\

DiffETM (ours)	& \textbf{0.2003} & \textbf{0.7504} & \textbf{0.1503}  & \textbf{547.1} & \textbf{0.1938} & \textbf{0.5940} &	\textbf{0.1151} & \textbf{470.7} & \textbf{0.1927 } & \textbf{0.2752} & \textbf{0.0530} & \textbf{596.6}     \\
    \bottomrule
    \end{tabular}}
    \linespread{1}
    \caption{This table presents the performance of our model and baselines on the 20NewsGroup dataset. K is the number of topics. The best results are bolded.}
    \label{tab:results_20}
\end{table*}

\linespread{1.25}
\begin{table*}[!ht]
    \centering
    \resizebox{0.9\textwidth}{20mm}{
    \begin{tabular}{c|cccc|cccc|cccc}
    \toprule
    \multirow{3}{*}{METHODS} & \multicolumn{12}{c}{Datasets} \\
    \cline{2-13} 
    & \multicolumn{4}{c|}{NYT-10000} & \multicolumn{4}{c|}{NYT-5000} & \multicolumn{4}{c}{NYT-3000}   \\
	& Coherence &	Diversity &  Quality & Perplexity & Coherence &	Diversity &  Quality & Perplexity  &  Coherence &	Diversity &  Quality & Perplexity  	\\

\midrule
NTMR & 0.078 & 0.3952 & 0.0307 & 881.4 & 0.0711 & 0.4224 & 0.0300 & 1358.3 & 0.0929 & 0.5504 & 0.0511 & 1734.0    \\
NTM	&	0.1811 & 0.4200	&	0.0761 & 679.0	&	0.1924 & 0.5552	&	0.1068 & 1066.1	& 0.2011 & 0.6064	&	0.1219 & 1377.7 \\
ETM	&	0.1885 & 0.6224	& 0.1173 & 642.1	&	0.2003 & 0.6416	&	0.1285 & 1064.7	&	0.2083 & 0.6704	&	0.1397 & 1372.7      \\
ERNTM	&	0.1888 & 0.6256	&	0.1181 & 644.1	&	0.2104 & 0.6768	&	0.1424 & 1060.2	&	0.2157 & 0.7096	&	0.1531 & 1365.9\\

DeTiME	&	0.1873 & 0.6645	& 0.1245 & 623.8	&	0.2116 & 0.6976 & 0.1476 & 1010.3	& 0.2173 & 0.7324	& 0.1592 & 1325.7     \\

Meta-CETM	&	0.1888 & 0.5947	& 0.1123 & 608.4	&	0.2122 & 0.6077 & 0.1290 & 1001.2	& 0.2187 & 0.5935	& 0.1298 & 1311.4     \\

DiffETM (ours)	&	\textbf{0.1906} & \textbf{0.7416} &  \textbf{0.1413} & \textbf{593.7} & \textbf{0.2145} & \textbf{0.7944} & \textbf{0.1704} & \textbf{996.2} & \textbf{0.2240} & \textbf{0.7704} & \textbf{0.1725} & \textbf{1304.6}      \\
    \bottomrule
    \end{tabular}}
    \linespread{1}
    \caption{This table presents the performance of our model and baselines on NewYorkTimes datasets. The topic number is set to 50. The best results are bolded.}
    \label{tab:results_nyt}
\end{table*}

\begin{table*}[!ht]
    \centering
    \resizebox{0.9\textwidth}{12mm}{
    \begin{tabular}{c|cccc|cccc|cccc}
    \hline
    \multirow{3}{*}{METHODS} & \multicolumn{12}{c}{Topic Number} \\
    \cline{2-13} 
    & \multicolumn{4}{c|}{K=50} & \multicolumn{4}{c|}{K=100} & \multicolumn{4}{c}{K=200}   \\
	& Coherence &	Diversity &  Quality & Perplexity & Coherence &	Diversity &  Quality & Perplexity  &  Coherence &	Diversity &  Quality & Perplexity  	\\
\midrule

DiffETM (ours)	& \textbf{0.2003} & \textbf{0.7504} & \textbf{0.1503}  & \textbf{547.1} & \textbf{0.1938} & \textbf{0.5940} &	\textbf{0.1151} & \textbf{470.7} & \textbf{0.1927 } & \textbf{0.2752} & \textbf{0.0530} & \textbf{596.6}     \\

-Diffusion	&	0.1945 & 0.7245	& 0.1409 & 788.4	&	0.1891 & 0.5266	& 0.0996 & 765.3	&	0.1875 & 0.2546 	& 0.0477 & 791.7 \\

ETM	&	0.1865 & 0.4864	&	0.0907 & 686.0	&	0.1821 & 0.3552	&	0.0647 & 660.0	&	0.1826 &  0.2326	&	0.0425 & 681.0\\
    \bottomrule
    \end{tabular}}
    \linespread{1}
    \caption{Ablation Study on 20NewsGroup datasets. The best results are bolded.}
    \label{ablation_20}
\end{table*}

\section{Experiment}

In this section, we assess the performance of our proposed method in comparison with embedded topic models (ETM~\cite{dieng2020topic}, ERNTM~\cite{Shao2022TowardsBU}), classic neural topic models (NTM~\cite{ding-etal-2018-coherence}, NTMR~\cite{ding-etal-2018-coherence}) and recent neural topic models (DeTiME~\cite{xu-etal-2023-detime}, Meta-CETM~\cite{NEURIPS2023_fce17645}).

We adopt the same hyper-parameters as NTM for fair comparisons. The parameters are initialized randomly for all experiments with the same random seed. We report topic coherence, topic diversity, and topic quality (the product between topic coherence and topic diversity) and document perplexity for each model on 20NewsGroup and NewYorkTimes datasets. All the codes were implemented using PyTorch~\cite{paszke2019pytorch}.

\subsection{Dataset's Statistical Information}
\label{datasets}
We conduct experiments on 20NewsGroup and NewYorkTimes datasets. The statistical information of datasets we used in this paper is shown in Table ~\ref{tab:data}.

\subsection{Implement Details}
\label{Implements}
In our settings, the word embedding size and topic embedding size are both 300 and we train these two embeddings in the training process. The diffusion step $T$ is 100, the $\beta_0$ is 0 and $\beta_{T}$ is 0.02. Batch sizes on 20NewsGroup and NewYorkTimes datasets are 1000 and 512, respectively. The $\lambda$ in the loss is 1. The learning rates on the 20Newsgroup dataset are 0.008, 0.009, and 0.01 for K=50, K=100 and K=200. The learning rates on the NewYorkTimes dataset are 0.007, 0.007 and 0.008 for NYT-3000, NYT-5000 and NYT-10000.

The metrics include topic coherence, topic diversity, topic quality, and perplexity. The detailed computations can be found in~\cite{Shao2022TowardsBU}. All experiments were conducted in a Linux server with a Nvidia V100 GPU.

\subsection{Main Results}
Since the performance of the topic model is influenced by the number of topics and the vocabulary size, we evaluate the model under different settings to better illustrate its performance. Specifically, we test models with varying topic numbers on the 20NewsGroup dataset and with different vocabulary sizes on the NewYorkTimes dataset. The detailed results are presented in Tables ~\ref{tab:results_20} and ~\ref{tab:results_nyt}.


In detail, Table ~\ref{tab:results_20} shows the comparison results on the 20NewsGroup dataset with different topic numbers. According to this table, our model outperforms all baselines with various topic numbers. Compared to ETM, the improvements are up to 77.89\%, which is achieved on the topic quality with 100 topics.

For the NewYorkTimes dataset, we construct different vocabulary sizes by removing words whose frequency is lower than 3000, 5000, and 10000. As shown in Table ~\ref{tab:results_nyt}, our proposed method reaches the best topic quality and topic perplexity on all three New York Times datasets.

\subsection{Ablation Study}

In order to verify the validity of the diffusion process of our model, we conducted the following ablation study: we compare our model with a model that undergoes the following modifications: (-Diffusion) removing diffusion process and directly using the document representation to produce the hidden representation. The results are shown in the Table ~\ref{ablation_20}. We find that compared to DiffETM, the model with diffusion removed has a slight decrease in metrics such as topic diversity, and topic coherence, but still outperforms ETM. The decline in perplexity metrics is large. The variant's performance on perplexity is weaker than ETM. These results indicate that the introduction of the diffusion process plays an important role in ensuring the smooth optimization of the model.

\subsection{Hyper-Parameter Analysis}
We observe that DiffETM has an important hyperparameter: the diffusion step T. This parameter determines the closeness of the hidden representations to the normal distribution, affecting the optimization process of the model as well as the final perplexity. To better illustrate the effect of this hyper-parameter on the model, we list the performance of the DiffETM on the 20NewsGroup dataset for different T at K = 50. The results are shown in Table ~\ref{hyp_T}. Based on the results in the table, we find that T has a large impact on perplexity. When T is small, perplexity is large. It indicates that the model is not well optimized. When T is large, perplexity is small and maintained at a stable stage. In addition, as T changes, the other two indicators also fluctuate slightly. In practice, we need to choose an appropriate T to keep the model harmonized on the three indicators.

\begin{table}[!ht]
    \centering
    \resizebox{0.7\linewidth}{12mm}{
    \begin{tabular}{c|cccc}
    \hline
T	& Coherence &	Diversity &  Quality & Perplexity \\
\midrule
0	&	0.1945 & 0.7245	& 0.1409 & 788.4	\\
20	&	0.1967 & 0.7456	& 0.1467 & 595.6	 \\
50	&	0.1992 & \textbf{0.7521}	& 0.1498 & 568.2	 \\
100	& \textbf{0.2003} & 0.7504 & \textbf{0.1503}  & 547.1 \\
150	&	0.1981 & 0.7021	& 0.1391 & 544.3	 \\
200	&	0.1959 & 0.6867	& 0.1345 & \textbf{542.6}	 \\
    \bottomrule
    \end{tabular}}
    \linespread{1}
    \caption{Performance of DiffETM with different T values on 20NewsGroup datasets. K is set to 50.}
    \label{hyp_T}
\end{table}


%% file: sections/conclusion.tex
\section{Conclusion}
After considering the limitations of the ETM in modeling document-topic distribution, we introduce the diffusion process into the sampling process of this distribution in the ETM for stronger modeling ability and to maintain a smooth optimization process. Extensive experiments conducted on two mainstream datasets have demonstrated our model's effectiveness in improving topic modeling performance.

%% file: main.bbl
\begin{thebibliography}{10}
\providecommand{\url}[1]{#1}
\csname url@samestyle\endcsname
\providecommand{\newblock}{\relax}
\providecommand{\bibinfo}[2]{#2}
\providecommand{\BIBentrySTDinterwordspacing}{\spaceskip=0pt\relax}
\providecommand{\BIBentryALTinterwordstretchfactor}{4}
\providecommand{\BIBentryALTinterwordspacing}{\spaceskip=\fontdimen2\font plus
\BIBentryALTinterwordstretchfactor\fontdimen3\font minus \fontdimen4\font\relax}
\providecommand{\BIBforeignlanguage}[2]{{%
\expandafter\ifx\csname l@#1\endcsname\relax
\typeout{** WARNING: IEEEtran.bst: No hyphenation pattern has been}%
\typeout{** loaded for the language `#1'. Using the pattern for}%
\typeout{** the default language instead.}%
\else
\language=\csname l@#1\endcsname
\fi
#2}}
\providecommand{\BIBdecl}{\relax}
\BIBdecl

\bibitem{dieng2020topic}
A.~B. Dieng, F.~J. Ruiz, and D.~M. Blei, ``Topic modeling in embedding spaces,'' \emph{Transactions of the Association for Computational Linguistics}, vol.~8, pp. 439--453, 2020.

\bibitem{Zhang2020CombineTM}
P.~Zhang, S.~Wang, D.~Li, X.~Li, and Z.~Xu, ``Combine topic modeling with semantic embedding: Embedding enhanced topic model,'' \emph{IEEE Transactions on Knowledge and Data Engineering}, vol.~32, pp. 2322--2335, 2020.

\bibitem{Akash2022CoordinatedTM}
P.~S. Akash, J.~Huang, and K.~C.-C. Chang, ``Coordinated topic modeling,'' \emph{ArXiv}, vol. abs/2210.08559, 2022.

\bibitem{Shao2022TowardsBU}
W.~Shao, L.~Huang, S.~Liu, S.~Ma, and L.~Song, ``Towards better understanding with uniformity and explicit regularization of embeddings in embedding-based neural topic models,'' \emph{2022 International Joint Conference on Neural Networks (IJCNN)}, pp. 1--9, 2022.

\bibitem{Seifollahi2021AnET}
S.~Seifollahi, M.~Piccardi, and A.~Jolfaei, ``An embedding-based topic model for document classification,'' \emph{Transactions on Asian and Low-Resource Language Information Processing}, vol.~20, pp. 1 -- 13, 2021.

\bibitem{wu2023effective}
X.~Wu, X.~Dong, T.~Nguyen, and A.~T. Luu, ``Effective neural topic modeling with embedding clustering regularization,'' in \emph{International Conference on Machine Learning}.\hskip 1em plus 0.5em minus 0.4em\relax PMLR, 2023.

\bibitem{xu-etal-2023-detime}
W.~Xu, W.~Hu, F.~Wu, and S.~Sengamedu, ``{D}e{T}i{ME}: Diffusion-enhanced topic modeling using encoder-decoder based {LLM},'' in \emph{Findings of the Association for Computational Linguistics: EMNLP 2023}, H.~Bouamor, J.~Pino, and K.~Bali, Eds.\hskip 1em plus 0.5em minus 0.4em\relax Singapore: Association for Computational Linguistics, Dec. 2023, pp. 9040--9057.

\bibitem{chen-etal-2023-nonlinear}
H.~Chen, P.~Mao, Y.~Lu, and Y.~Rao, ``Nonlinear structural equation model guided {G}aussian mixture hierarchical topic modeling,'' in \emph{Proceedings of the 61st Annual Meeting of the Association for Computational Linguistics (Volume 1: Long Papers)}, A.~Rogers, J.~Boyd-Graber, and N.~Okazaki, Eds.\hskip 1em plus 0.5em minus 0.4em\relax Toronto, Canada: Association for Computational Linguistics, Jul. 2023, pp. 10\,377--10\,390.

\bibitem{NEURIPS2023_fce17645}
Y.~Xu, J.~Sun, Y.~Su, X.~Liu, Z.~Duan, B.~Chen, and M.~Zhou, ``Context-guided embedding adaptation for effective topic modeling in low-resource regimes,'' in \emph{Advances in Neural Information Processing Systems}, A.~Oh, T.~Neumann, A.~Globerson, K.~Saenko, M.~Hardt, and S.~Levine, Eds., vol.~36.\hskip 1em plus 0.5em minus 0.4em\relax Curran Associates, Inc., 2023, pp. 79\,959--79\,979.

\bibitem{Kingma2013AutoEncodingVB}
D.~P. Kingma and M.~Welling, ``Auto-encoding variational bayes,'' \emph{CoRR}, vol. abs/1312.6114, 2013.

\bibitem{miao2017discovering}
Y.~Miao, E.~Grefenstette, and P.~Blunsom, ``Discovering discrete latent topics with neural variational inference,'' in \emph{International Conference on Machine Learning}.\hskip 1em plus 0.5em minus 0.4em\relax PMLR, 2017, pp. 2410--2419.

\bibitem{bank2021autoencoders}
\BIBentryALTinterwordspacing
D.~Bank, N.~Koenigstein, and R.~Giryes, ``Autoencoders,'' 2021. [Online]. Available: \url{https://arxiv.org/abs/2003.05991}
\BIBentrySTDinterwordspacing

\bibitem{Nichol2021ImprovedDD}
A.~Nichol and P.~Dhariwal, ``Improved denoising diffusion probabilistic models,'' \emph{ArXiv}, vol. abs/2102.09672, 2021.

\bibitem{Newsgroups20}
\BIBentryALTinterwordspacing
{empty}, ``20 newsgroups dataset,'' {empty}. [Online]. Available: \url{http://people.csail.mit.edu/jrennie/20Newsgroups/}
\BIBentrySTDinterwordspacing

\bibitem{10.1007/s11042-018-6894-4}
\BIBentryALTinterwordspacing
H.~Jelodar, Y.~Wang, C.~Yuan, X.~Feng, X.~Jiang, Y.~Li, and L.~Zhao, ``Latent dirichlet allocation (lda) and topic modeling: models, applications, a survey,'' \emph{Multimedia Tools Appl.}, vol.~78, no.~11, p. 15169–15211, jun 2019. [Online]. Available: \url{https://doi.org/10.1007/s11042-018-6894-4}
\BIBentrySTDinterwordspacing

\bibitem{doi:10.1080/00437956.1954.11659520}
Z.~S. Harris, ``Distributional structure,'' \emph{WORD}, vol.~10, no. 2-3, pp. 146--162, 1954.

\bibitem{ding-etal-2018-coherence}
\BIBentryALTinterwordspacing
R.~Ding, R.~Nallapati, and B.~Xiang, ``Coherence-aware neural topic modeling,'' in \emph{Proceedings of the 2018 Conference on Empirical Methods in Natural Language Processing}.\hskip 1em plus 0.5em minus 0.4em\relax Brussels, Belgium: Association for Computational Linguistics, Oct.-Nov. 2018, pp. 830--836. [Online]. Available: \url{https://www.aclweb.org/anthology/D18-1096}
\BIBentrySTDinterwordspacing

\bibitem{paszke2019pytorch}
A.~Paszke, S.~Gross, F.~Massa, A.~Lerer, J.~Bradbury, G.~Chanan, T.~Killeen, Z.~Lin, N.~Gimelshein, L.~Antiga \emph{et~al.}, ``Pytorch: An imperative style, high-performance deep learning library,'' \emph{arXiv preprint arXiv:1912.01703}, 2019.

\end{thebibliography}
